\documentclass[letterpaper, 10 pt, conference]{ieeeconf}
\IEEEoverridecommandlockouts
\usepackage{cite}
\usepackage{amsmath,amssymb,amsfonts}
\usepackage{algorithmic}
\usepackage{graphicx}
\usepackage{textcomp}
\usepackage{xcolor}
\usepackage{balance}
\usepackage{float}
\def\BibTeX{{\rm B\kern-.05em{\sc i\kern-.025em b}\kern-.08em
   T\kern-.1667em\lower.7ex\hbox{E}\kern-.125emX}}

\newtheorem{definition}{Definition}

\begin{document}

\title{\LARGE \bf Active Exploration for Real-Time Haptic Training}

\author{Jake Ketchum$^{1}$, Ahalya Prabhakar$^{2}$, Todd D. Murphey$^{1}$
\thanks{$^{1}$Center for Robotics and Biosystems, Northwestern University, Evanston, IL, USA.}%
\thanks{$^{2}$Department of Mechanical Engineering and Materials Science, Yale University, New Haven, CT, USA.}
}


\maketitle

\begin{abstract}

Tactile perception is important for robotic systems that interact with the world through touch. Touch is an active sense in which tactile measurements depend on the contact properties of an interaction---e.g., velocity, force, acceleration---as well as properties of the sensor and object under test. These dependencies make training tactile perceptual models challenging. Additionally, the effects of limited sensor life and the near-field nature of tactile sensors preclude the practical collection of exhaustive data sets even for fairly simple objects. Active learning provides a mechanism for focusing on only the most informative 
aspects of an object during data collection. Here we employ an active learning approach that uses a data-driven model's entropy as an uncertainty measure and explore relative to that entropy conditioned on the sensor state variables. Using a coverage-based ergodic controller, we train perceptual models in near-real time. We demonstrate our approach using a biomimentic sensor, exploring ``tactile scenes" composed of shapes, textures, and objects. Each learned representation provides a perceptual sensor model for a particular tactile scene. Models trained on actively collected data outperform their randomly collected counterparts in real-time training tests. Additionally, we find that the resulting network entropy maps can be used to identify high salience portions of a tactile scene.

\end{abstract}

\section{Introduction}

Touch enables locating objects and navigating spaces without relying on vision. For instance, rummaging through a drawer or locating a light-switch in the dark are both parts of everyday life. The ability to operate in fully or partially vision denied environments, say inside a machine, is essential for a wide range of tasks in maintenance, inspection, and manufacturing. Over the last several decades, a number of sensors have been developed which provide human, or at least human-adjacent, levels of performance across a range of tactile sensing tasks. Despite significant work in the area, robust tactile sensing and perception is aspirational in most deployed robots today. 

\begin{figure}[h]
\centering
\includegraphics[width = 6.75cm]{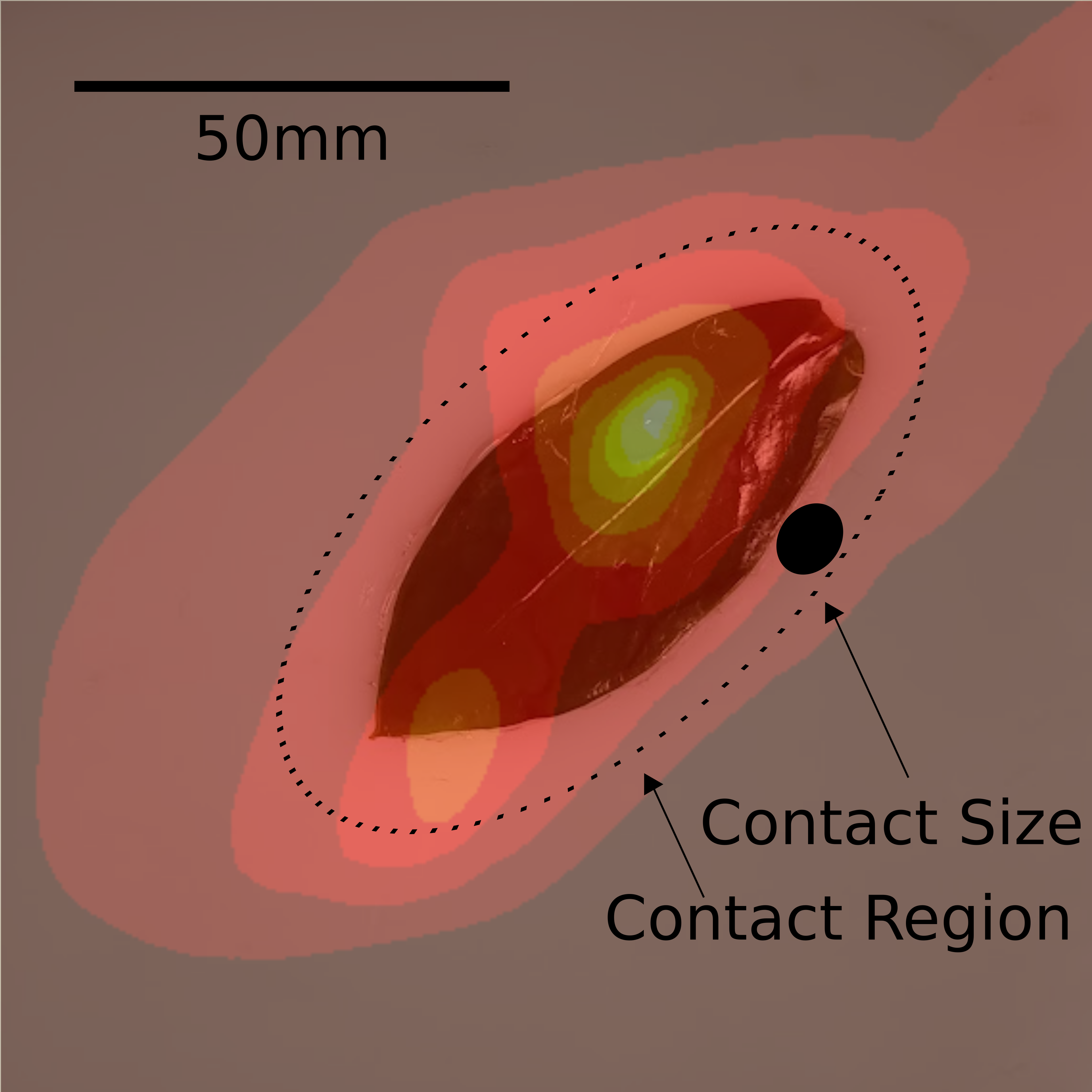}
\caption{ \textbf{Haptic exploration of a leaf on a test token:} The high entropy regions of the neural network indicate where the sensor should collect data in the scene---the image edge is the edge of the reachable scene---most relevant to predicting future measurement values. The black ellipse indicates the approximate size of the sensor footprint and the high information content areas include both when the sensor is in direct contact as well as when the sensor is adjacent to the leaf.}
\label{fig: contact_overlay}
\end{figure}

This gap between what we observe in tactile animal behavior and what we would like to observe in robotic systems is partially due to touch being an obligate active sense (in contrast to vision and aural senses that can be effective as passive sensors). 
The output of a sensor is dependent not only on the environment and object under test, but also on the contact conditions---e.g., relative velocity, acceleration, and pressure---of the sensor itself. Furthermore, biomimetic tactile sensors will, in almost all cases, be near-field and must explore an object extensively in order to collect information about its size and shape. Biomimetic tactile sensors are also frequently high dimensional, stochastic and non-linear, making manual processing of the data difficult. One solution to these challenges is to use data-driven (learned) perceptual models, that ingest raw data and return synthesized models of the tactile landscape. 

We present a method for synthesizing  exploratory behaviors to accelerate training of haptic perceptual models, using the SynTouch BioTac sensor as a model haptic sensor. Our primary contribution in this work is an active learning strategy for generating tactile exploratory motions by exploring relative to a generative model's conditional uncertainty distribution. This enables the sensor to spend more time in high information areas of the space, as determined by the underlying machine learning model. We demonstrate that this method produces more accurate perceptual models during real-time training as compared to a random exploration baseline across six different tactile scenes with a variety of textures and geometries. We moreover demonstrate that the uncertainty maps from these models can be used to identify areas of high importance in a tactile scene, even when those areas are caused by organic objects like leaves which defy easy manual definition. 

This paper is organized as follows: Section \ref{section: related work} discusses important background relating to the model architecture, sensor, and learning strategy employed in this paper. Section \ref{section: method} outlines the theoretical framework and software architecture used for our experiments. Section \ref{section: Hardware} provides details about experimental hardware. Finally in section \ref{section: Results} we discuss the performance of our method. 

\section{Related Work}
\label{section: related work}

This work relies on active learning and generative neural networks, in particular autoencoders. We also make use of the SynTouch BioTac to demonstrate our method. The following subsections provide additional background. 

\subsection{CVAEs}
Autoencoders (AEs) are a neural network variant which find lower dimensional data representations in a manner analogous to non-linear principal component analysis. Autoencoders are typically composed of two elements: an encoder network and a decoder network. The encoder network takes full dimension input data and returns a lower dimensional latent representation. The decoder network takes an element of the latent space and returns a signal in the same space as the original input; training is based on requiring the decoder output to approximate the original input data. The encoder network can be used as a preprocesser for efficiently training other models to perform tasks such as classification or control. 

To ensure better out-of-distribution performance, variational autoencoders (VAEs) modify the encoder to produce a multivariate latent distribution. This distribution is then sampled to provide a latent vector for the decoder network \cite{kingma_auto-encoding_2022}. The latent distribution acts as a regularizer, and improves performance in regions of the latent space with sparse training data \cite{kumar_implicit_2020}. 

A final modification of the AE structure introduces a conditional vector associated with each data point. The conditional vector is provided as an input to both constituent networks as shown in figure \ref{fig: cvae}. During inference the conditional input to the decoder can be varied to predict how sensor readings would change given different conditional parameters, so that the decoder can be used as a generative model. The resulting model is called a Conditional Variational Autoencoder (CVAE) \cite{sohn_learning_2015}. 

Unsupervised training makes VAEs particularly promising for use in settings where novelty is expected (e.g., infrastructure monitoring \cite{zhang_unsupervised_2022}). VAEs and CVAEs have also been used for sensor modeling, including for compressing radar images and modeling soft sensors \cite{zhu_parallel_2022, wang_cvae_2023, dixit_intelligent_2020}. CVAEs are particularly promising for perceptual modeling, because the conditional vector provides a way to extract predicted sensor data across a variety of sensor states \cite{wang_data_2020, itkina_multi-agent_2022}, making them particularly appropriate for haptic perception.

\subsection{Active Learning}
Active learning (AL) is a machine learning strategy in which information about the model state or performance is used to guide data collection. This can result in lower losses or greater data efficiency relative to conventional training strategies. In deep learning AL can improve training times by biasing the data set, or reduce data labeling costs by prioritizing certain samples for manual review \cite{cohn_active_1996}. In a robotics context AL can be used to guide the behavior of a system to maximize model learning. For example, an active learning approach increases the rate at which Koopman operators can be re-trained on simulated quadroter dynamics in flight \cite{abraham_active_2019}.
AL can additionally be powerful for training perceptual sensor models, since it enables a data collection system to focus on only the most important parts of a sensor target. 

Training speed, data efficiency, and energy use are all key considerations when training learned sensor models. Active learning can be used to substantially improve training performance of CVAEs on camera-like sensors \cite{prabhakar_mechanical_2022}. However, active learning has not been used for generating perceptual models of tactile sensors. Since tactile sensors tend to be delicate and slow to collect samples, they represent a particularly promising application area of AL, motivating the present work. 

\subsection{BioTac Sensor}
We demonstrate our methods using a BioTac sensor from SynTouch. The BioTac is a biomimetic finger-tip sensor designed to have sensing capabilities similar to those of a human finger \cite{lin_signal_2009}. The BioTac has three primary sensing modalities: an ultrasonic pressure sensor which provides spectral data, an array of 19 electrodes which provide spatial pressure data, and a temperature sensor which provides a measurement of heat-flux\cite{lin_signal_2009}. The BioTac is composed of a solid inner core, containing the sensing electronics, a soft conductive skin, and an electrolytic fluid layer\cite{lin_signal_2009}. The outer skin contains ``fingerprint" ridges which help the ultrasonic sensor resolve textures, and it is held in place by a rigid ``fingernail" which also serves to seal the fluid chamber \cite{lin_signal_2009}. 

Early research using the BioTac showed near-human microvibration and impact detection performance using the ultrasonic sensor\cite{fishel_sensing_2012}. In \cite{fishel_bayesian_2012} it was demonstrated that a set of 3 features generated from the ultrasonic data can be used to accurately identify various textures using Bayesian exploration. Recently, a spiking network was applied for rapid texture identification using the BioTac's electrode array across 20 different materials \cite{taunyazov_fast_2020}. The BioTac is extensively used in haptic robotic manipulation, including for haptic teleoperation and manipulation of diverse objects \cite{reinecke_experimental_2014, liang_-hand_2020, fishel_tactile_2020, pacchierotti_cutaneous_2016}. 

Although the BioTac sensor provides a rich and sensitive data stream, it experiences drift and is non-linear in force response beyond 2N \cite{fishel_bayesian_2012}. Finite Element Analysis (FEA) simulations and data sets have improved modeling of the BioTac sensor \cite{narang_interpreting_2021, ruppel_simulation_2019, chebotar_bigs_2016}. However, while there are potentially adequate simulation capabilities for a BioTac in a manipulation context, there are no simulations suitable for predicting BioTac texture response. 

\section{Method}
\label{section: method}
Our method efficiently trains tactile sensor models for processing tactile data to enable higher level robotic reasoning (e.g., object identification, registration, and localization). We have selected a CVAE for the model architecture and an ergodic exploration for generating exploratory motions. We demonstrate this approach using a robotic gantry, a SynTouch Biotac sensor, and a number of 6"x6" tactile scenes.

\subsection{Learning Architecture} 
\begin{figure}[h]
\includegraphics[width = 8.75cm]{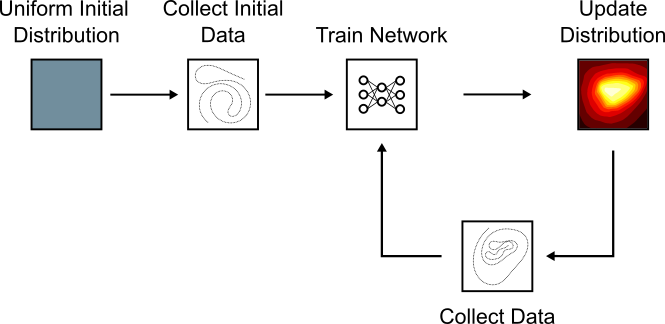}
\caption{\textbf{Closed loop data collection for haptic learning: }
After initially collecting data using a uniform distribution as the specification, the system collects data based on the state of the network while simultaneously training the network based on collected data.
}
\label{fig: closed loop learning}
\end{figure}

Our learning process (Fig.~\ref{fig: closed loop learning}) alternates between periods in which the model is trained using previously collected data, and periods in which the state of the model is used to guide the collection of new data. 


\subsection{Model Architecture}

\begin{figure}[h]
\includegraphics[width = 8.75cm]{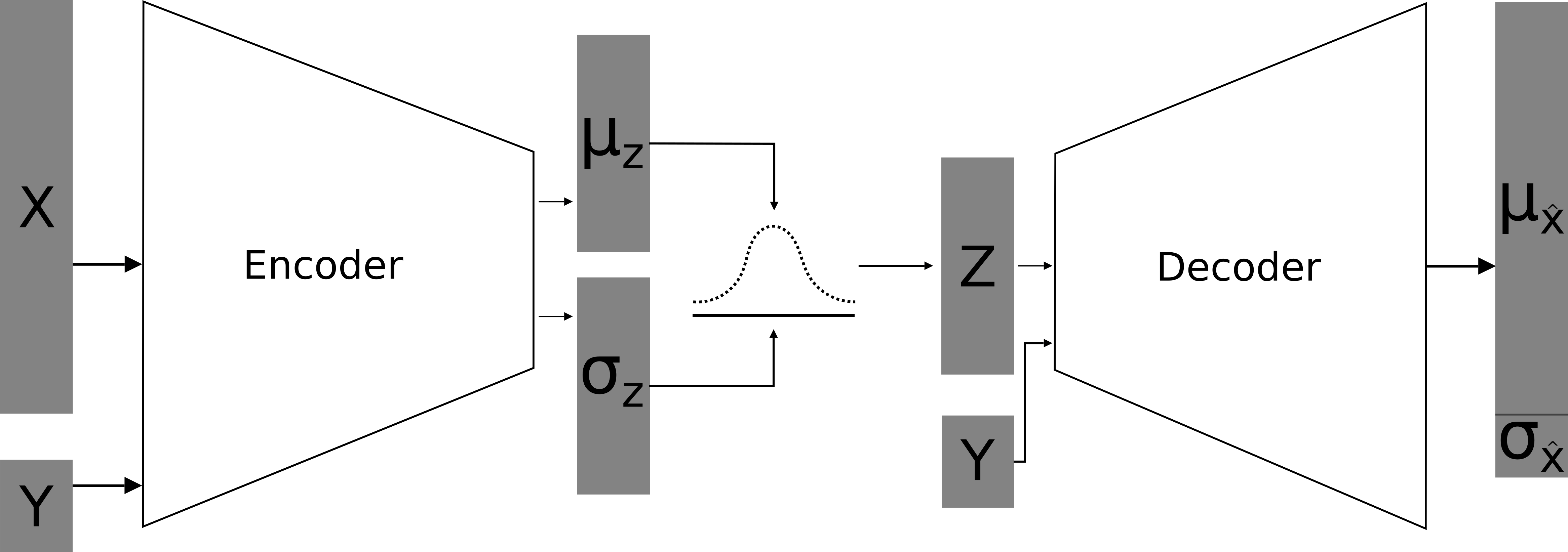}
\caption{\textbf{Network architecture: } The haptic measurement is conditioned on the sensor state while the decoder predicts the measurement at any sensor state.
}
\label{fig: cvae}
\end{figure} 

Our work uses the CVAE architecture shown in Fig.~\ref{fig: cvae}. This model architecture was selected for its ability to efficiently compress structured data, and because it can be used to make predictions about sensor values for different system states. The model's data input has 57 dimensions consisting of three concatenated 19 dimensional sensor readings, and its conditional input is the position of the sensor in Cartesian space. The model's output distribution is a maximum likelihood Gaussian modeling the encoder input. The output distribution has a scalar diagonal variance matrix with a magnitude provided by the decoder. This variance output from the decoder provides a measurement of the network's uncertainty for a particular pairing of a latent space element and conditional vector, which might reflect either underlying variance in the data or the training state of the network.

The encoder and decoder networks are composed of fully connected layers and taper symmetrically towards the center of the network. Each half of the network has an inter-layer ratio of $0.8$, with an initial layer width of $300$, and a depth of $4$ layers. The encoder outputs means and variances for a $6$ dimensional latent distribution with a diagonal variance matrix. During training a value is sampled from this distribution and passed to the decoder as the model's latent vector. The model is trained with a batch size of $256$ and a dropout value of $p = 0.2$. Leaky ReLU activation functions are used for all neurons except the output layer, which uses a signmoid function to enforce consistent scaling. The network is initialized with Kaiming Uniform Initialization for $a = sqrt(5)$\cite{he_delving_2015}. 

\subsection{Ergodic Metric}

\begin{definition}[Ergodic]
An agent's trajectory is ergodic with respect to some distribution, if the trajectory's time-averaged statistics are equal to the distribution's spatial statistics. 
\end{definition}


\begin{figure*}
\centering
\includegraphics[width = 10.75cm]{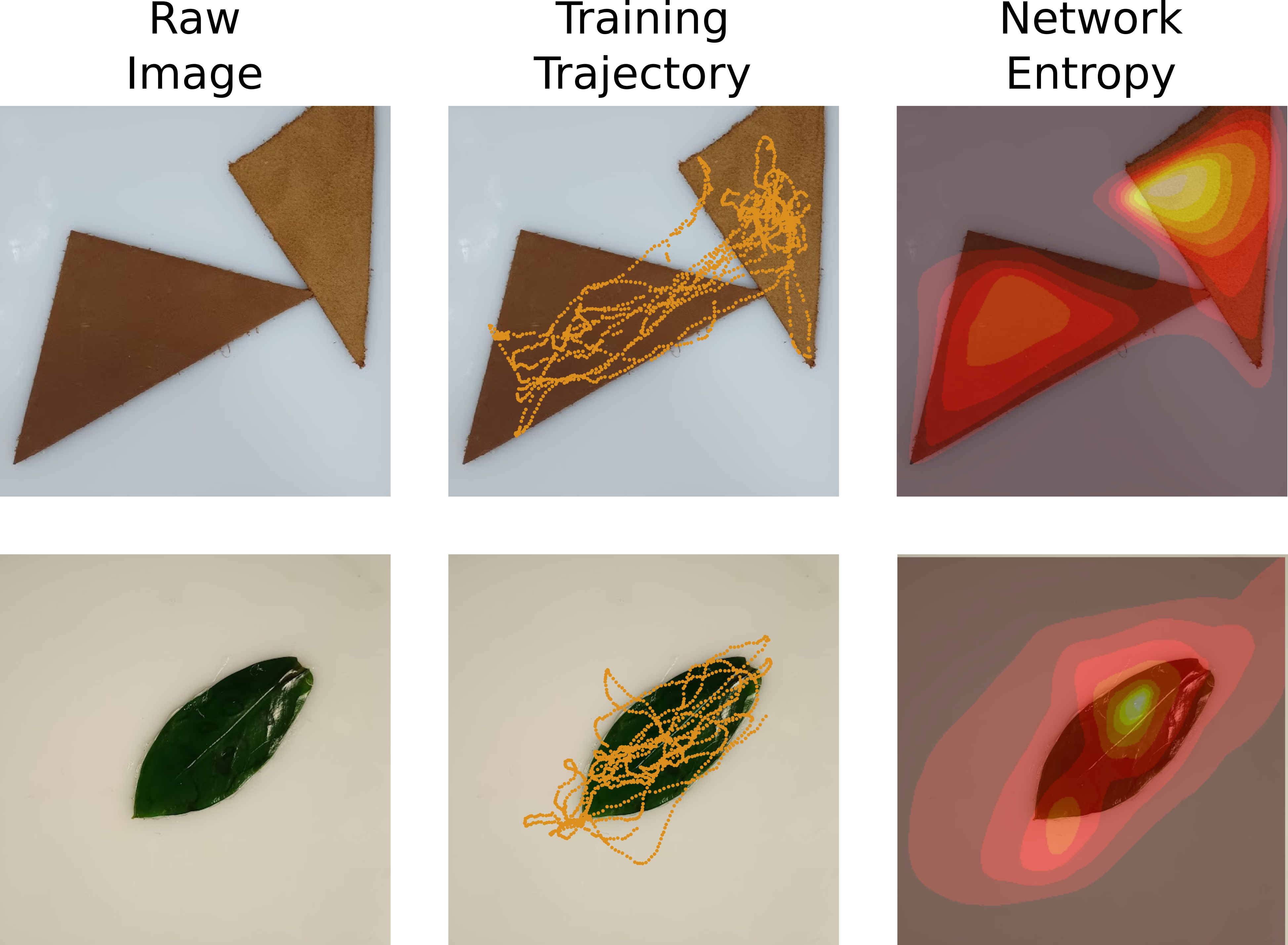}
\caption{\textbf{Sample scenes, exploration trajectory, network entropy:} For each data collection episode a sensor exploration trajectory is generated using the network entropy as a target distribution. This process ensures that the sensor spends time in informative areas of the scene. }
\label{fig: leaf-n-figure}
\end{figure*}

Generating exploratory motions for tactile sensing involves balancing two competing priorities: The sensor must explore high-salience parts of a scene under many contact conditions in order to build a repeatable model, but it must also continue to explore other parts of the environment and other contact conditions in case they contain high salience features. The coverage of a sensor trajectory relative to the environmental domain and contact conditions can be specified by minimizing the ergodic metric \cite{mathew_metrics_2011} relative to a continuous target distribution provided by a learning model. 

In order to compare a trajectory (a set of states parameterized by one continuous variable time $t$) to a 2D (or higher dimensional) distribution, the trajectory is first decomposed into a series of delta functions, Eq.~\eqref{equ: traj delta}. 

    \begin{equation}
    \label{equ: traj delta}
    C(x) = \frac{1}{t} \int_{0}^{t} \delta(x - x(\tau)) d\tau
    \end{equation}

These delta functions can be represented using a Fourier decomposition of \eqref{equ: traj delta} using basis functions of the form \eqref{equ: fourier basis} where $n$ is the dimension of the state space, $k$ is an index over the coefficients of the multidimensional Fourier transform, $L_i$ is the length of the $i^{th}$ dimension and $h_k$ is a normalizing factor to ensure the basis remains orthogonal. 

    \begin{equation}
        \label{equ: fourier basis}
        F_k(x(t)) = \frac{1}{h_k}\prod_{i=1}^{n}\cos(\frac{k_1\pi}{L_i}x_i(t))
    \end{equation}

    \begin{equation}
        \label{equ: 2D normalizer}
        h_k = (\int_0^{L_1} \int_0^{L_2} \cos^2(\frac{k_1\pi}{L_1}x_1)\cos^2(\frac{k_2\pi}{L_2}x_2)dx_1dx_2)^{\frac{1}{2}}
    \end{equation}

The Fourier coefficients corresponding to the trajectory are given by \eqref{equ: traj coefficients} and the coefficients approximating the target distribution are given by \eqref{equ: target coefficients}. 

    \begin{equation}
        \label{equ: traj coefficients}
        c_k = \frac{1}{T} \int_0^T F_k(x(t))dt
    \end{equation}
    \begin{equation}
        \label{equ: target coefficients}
        \phi_k = \int_X \phi(x)F_k(x)dx
    \end{equation}

The ergodic metric can then be computed by taking the (Sobolev) distance between the time-averaged trajectory statistics and the target distribution \eqref{equ: ergodic metric}.

    \begin{equation}
        \label{equ: ergodic metric}
        \epsilon(t) = \sum_{k_1=0}^{K} ... \sum_{k_n=0}^{K}(1+||k||^2)^{-\frac{n+1}{2}}|c_k-\phi_k|^2
    \end{equation}
    
For faster computation, we use a version of the sample-based ergodic controller described in \cite{abraham_ergodic_2021}. This formulation approximates the ergodic measure based on Kullback-Leibler divergence and replaces the Fourier based ergodic metric above with \eqref{equ: kldiv objective}. $P(s)$ is the target distribution, $q(s) = q(s | x(t))$, and s represents a sample. 

    \begin{equation}
        \label{equ: kldiv objective}
        D_{KL}(p||q) = -E_{p(s)}[\log(q(s))]
    \end{equation}

    The expectation can then be replaced with a sample based approximation as shown in \eqref{equ: kl final}. In our case, the distribution being sampled is provided by the model under training. This allows the target distribution to evolve as the model learns and new areas of high tactile complexity are discovered. 

    \begin{equation}
        \label{equ: kl final}
        D_{KL}(p||q) \approx - \sum_{i = 1}^N P(s_i)\log(q(s_i))
    \end{equation}
Haptic exploration benefits from this specification by always representing the entire domain and the likelihood that a sensor reading anywhere in the domain will improve the generative predictions for the haptic sensor. As a result, the exploration strategy can always take into account the entire domain \emph{and} the state of the learning model over that whole domain, avoiding fixating on a small subset of the domain that happens to be feature rich.

\subsection{Exploratory Motions}

The premise of our approach to generating exploratory motions is that by spending more time in areas of the conditional space with high model entropy, the sensor will collect a higher quality data set. Since the model output is structured as a multivariate Gaussian, \eqref{equ: general guassian entropy}---which describes the entropy of a multivariate Gaussian---can be used to calculate the model entropy for a given decoder input. For \eqref{equ: general guassian entropy}: $x \sim N_D(\mu, \Sigma)$ is the network output distribution, $Y$ is the conditional vector, $Z$ is the latent vector, and $D = 57$ is the dimension of a data point.

\begin{equation}
    \label{equ: general guassian entropy}
    H(Y, Z) = \frac{D}{2}(1+log(2\pi)) + \frac{1}{2}log(|\Sigma|)
\end{equation}

Since for our model the output variance ($\Sigma = \sigma I_{D}$) is a scalar matrix, \eqref{equ: general guassian entropy} simplifies to \eqref{equ: guassian simplified}. 

\begin{equation}
    \label{equ: guassian simplified}
    H(Y, Z) = \frac{D}{2}(1+log(2\pi)) +\frac{D}{2}log(\sigma)
\end{equation}

This value can then be sampled for different conditional vectors to provide the target distribution required for ergodic exploration. A fixed length trajectory is then optimized relative to this target distribution using the ergodic metric described by \eqref{equ: kl final}. As an example of this process, figure \ref{fig: leaf-n-figure} shows two target distributions (Network Entropy) and the exploratory trajectories that result (Training Trajectory). In practice, each time the distribution is sampled we use several latent vectors drawn from the preceding round of data collection and then average the result. This ensures a consistent distribution during the trajectory generation process.

To encourage exploration we add a constant value to all sampled entropies, ensuring that at least $5\%$ of the sensor's time is spent in low-entropy areas of the conditional space. This helps prevent the neural network from fixating on high-entropy areas.

The initial data collection round---prior to any training---needs to be specified. To encourage exploration the initial target distribution is set to a uniform distribution across the conditional space. This ensures the sensor explores the entire scene before beginning the active learning process. 

For comparison purposes, we use a passive sampling approach that does not use the state of the learning model, specifically relying on random walks, though raster scanning could also be used. Random walks can be vulnerable to over-exploring low information areas of the space, or to missing key areas---like a leaf, piece of leather, or another ``object"---entirely. Rastering strategies ensure good coverage, but can take a long time to achieve good coverage and risk being dominated by irrelevant data. This is particularly an issue for tactile sensors, which must explore not only a range of positions/states, but also a range of contact parameters.

\section{Hardware}
\label{section: Hardware}

This work uses a SynTouch BioTac sensor mounted to a modified X-Carve gantry. Position tracking is provided by April Tags and a Logitech Brio. To ensure accuracy, the vision system and gantry are calibrated against each other at the start of each training run. The gantry speed is capped at $120\frac{mm}{s}$ in order to minimize tracking error. All of the experiments in this paper were run using a desktop with a Ryzen 2970WX CPU and a GTX2080 GPU. 

\begin{figure}[h]
\includegraphics[width = 8.75cm]{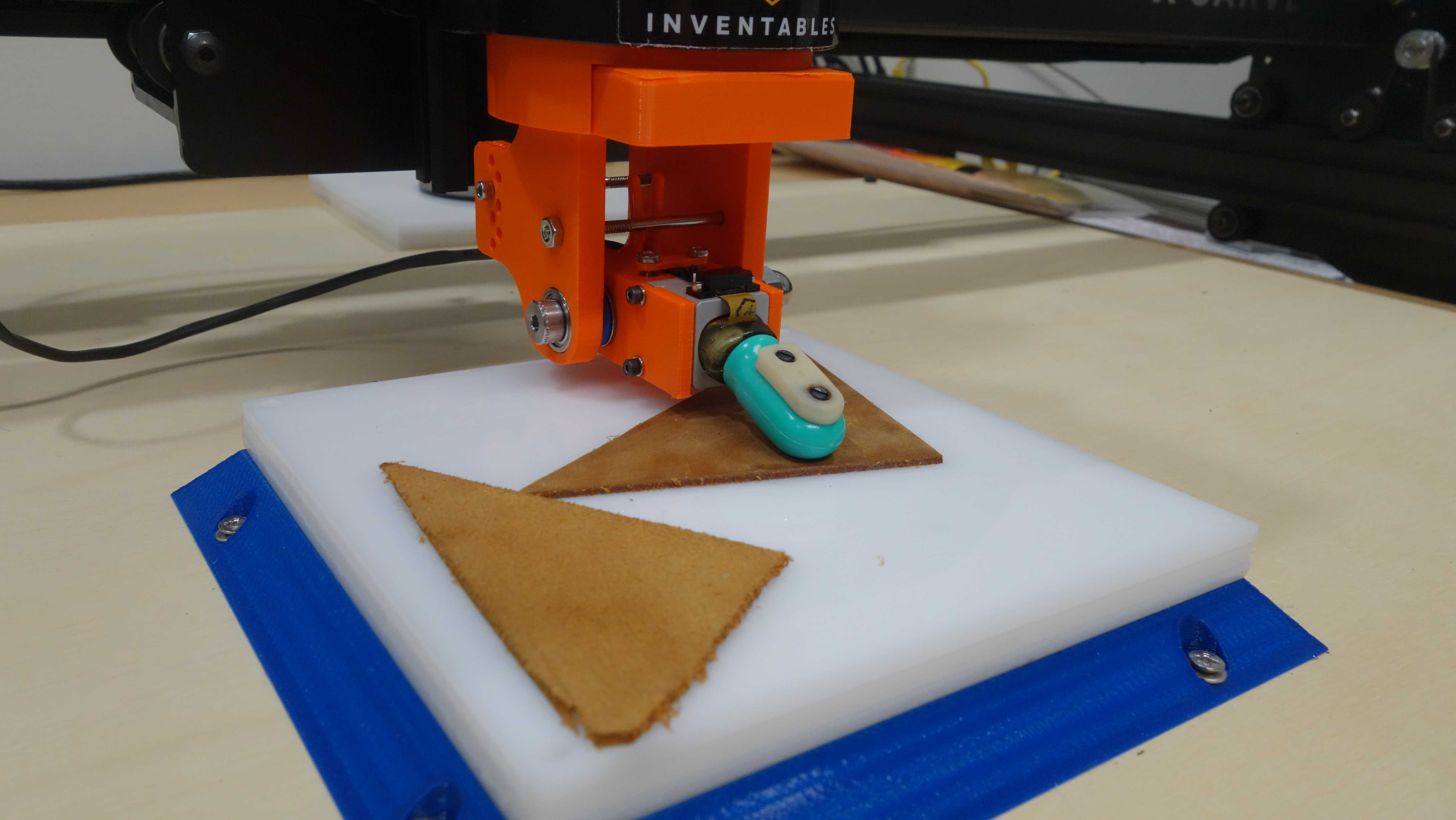}
\caption{\textbf{Experimental gantry, BioTac sensor and haptic tokens:}
The interchangeable token each constitute a tactile scene and are held with pins for hot swapping. The sensor's compliant mount can accommodate 6mm of height variation.
}
\label{fig: biotac head}
\end{figure}

Six different tactile scenes were used in this work including various shapes cut from leather and acrylic, arrangements of painters and duct tape, and the leaves of a Zanzibar Gem plant. Each sample was arranged onto a 145mm x 145mm acrylic sample token as show in figure \ref{fig: biotac head}. These materials were chosen to represent a range of non-abrasive textures, and were arranged into various shapes to create ``tactile scenes". Where necessary samples were flattened and adhered to their acrylic carriers with double-stick tape, hot glue or CA glue. 

The BioTac sensor is mounted on a pivoting compliant base with a preload of $1.25N$ and a spring constant of $0.39\frac{N}{mm}$. Since the BioTac sensor is known to exhibit thermal drift, the contact force was re-calibrated prior to each data collection run. To further mitigate thermal drift the sensor was also allowed to run for at least an hour prior to use. However, we found that even with these measures the data varied substantially from day to day, and even between different collection runs on the same day. To protect the sensor's skin from wear the pinkie finger of a small nitrile glove was used as an outer covering. This covering was replaced whenever any sign of wear occurred, similar to  \cite{fishel_bayesian_2012}. 

Data was collected from the BioTac at a rate of 100hz, and positions were recorded by the camera at a rate of 30hz. Only the BioTac’s 19 spatial electrodes were used---data from the heat flow and ultrasonic sensors were discarded. A single data point for training consisted of one position reading and the three preceding BioTac readings---resulting in 2 conditional dimensions and 57 electrode dimensions. Prior to use each electrode value is normalized to a range of [0,1]. 

\section{Results}
\label{section: Results}

\begin{figure}[h]
\includegraphics[width = 8.75cm]{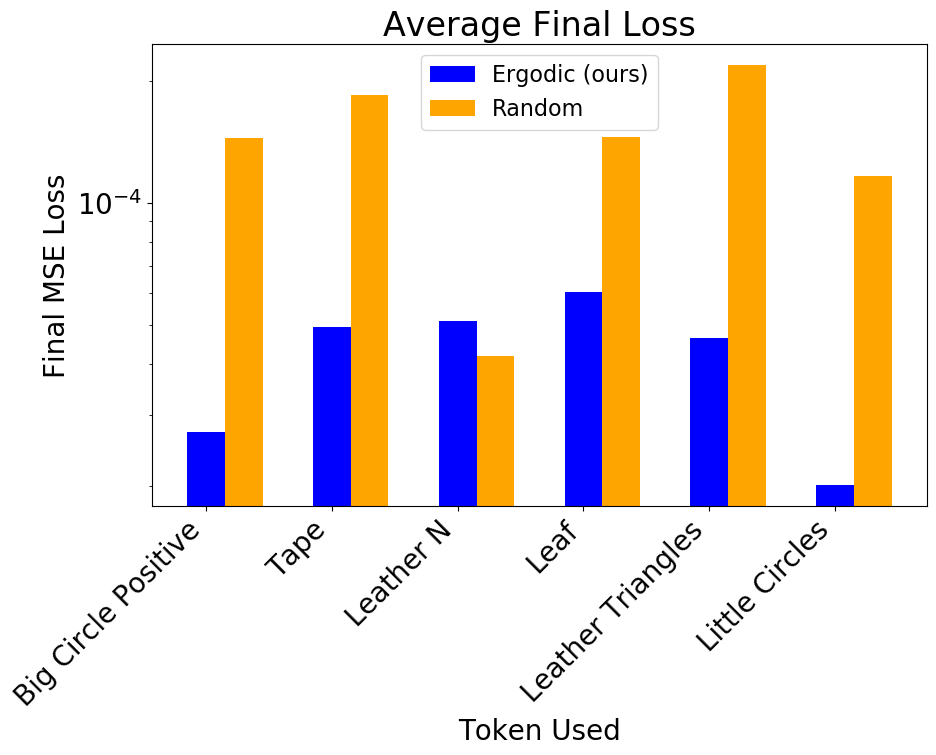}
\caption{\textbf{Average MSE losses for real-time training on 6 different tactile scenes:} Note that in all but one case the active ergodic exploration substantially outperformed the random walk. In the one case where the two are comparable, the Leather `N', almost the entire scene has high entropy.}
\label{fig: online bar chart}
\end{figure}

We tested our method using six different tactile scenes as described in section \ref{section: method}. We selected these scenes to span a range of conditions including partial contact, multiple textures, complex shapes, and organic materials. For each scene we conducted two training runs using a random walk data collection strategy, and two runs using our active learning method. To provide a consistent comparison, we collected the same total path-length of data for all trials. As shown in figure \ref{fig: online bar chart} we found that the active learning method outperformed the baseline in every run on five of the six scenes. For the leather `N', the random walk was expected to be comparable to the active learning approach since the high information areas cover almost the entire scene.

\begin{figure}[h]
\centering
\includegraphics[width=7.5cm]{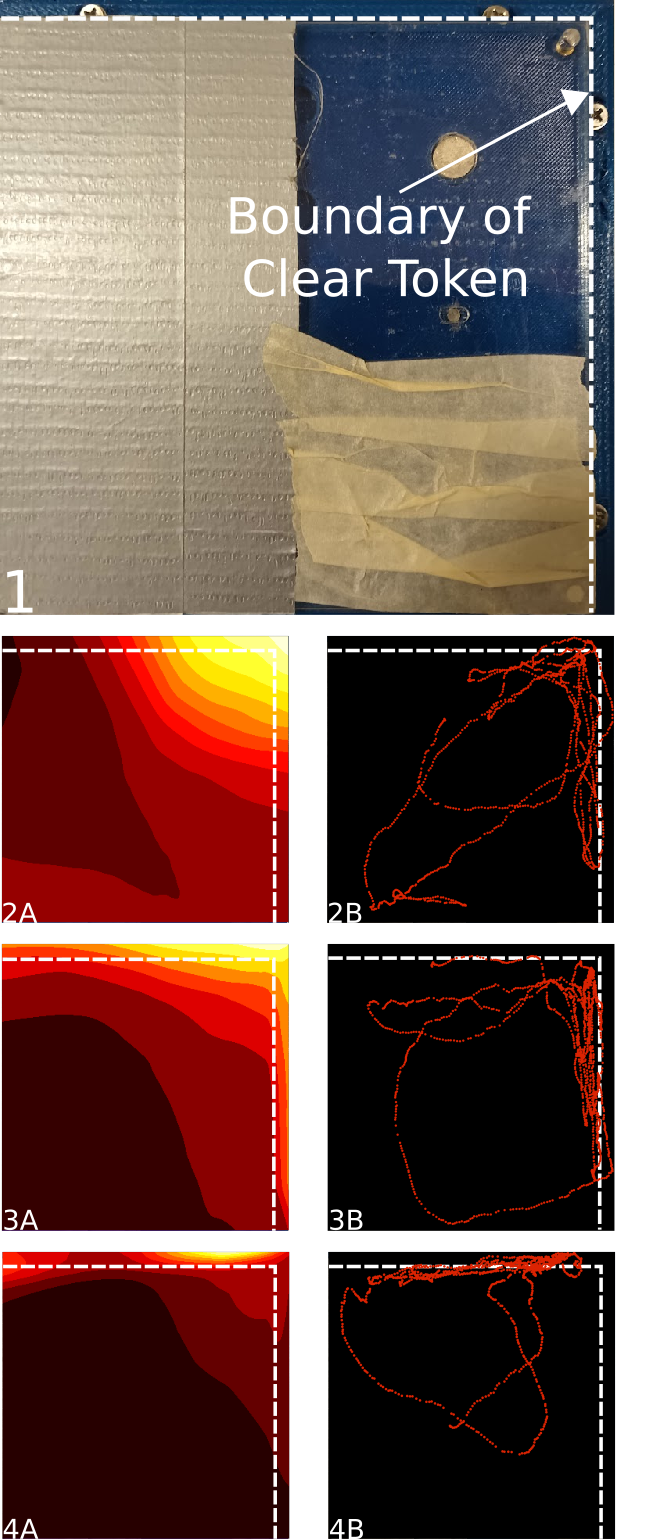}
\caption{\textbf{Haptic exploration naturally focuses on edges and corners:} 
When given access to the token edge (1), the model clearly identified the exposed edges and corner as high salience areas. The heatmaps (column A) show how the focus shifted along the boundary during different collection episodes. The trajectories in column B show the sensor movement for each episode.}
\label{fig: top edge entropy}
\end{figure}

Edges and corners are important haptic features, and during training the active exploration intermittently fixates on both edges and corners (Fig.~\ref{fig: top edge entropy}). This behavior was most prevalent during a series of trials in which the system was able to access the upper edge of one of the tokens. These tests consistently produced entropy distributions similar to the one in figure \ref{fig: top edge entropy}. That distribution clearly highlights the upper edge, right edge, and upper right corner of the tactile token. Moreover, the experimental response to detecting these edges and corners leads to the exploratory motion concentrating time on modeling them. 

The natural emergence of this behavior is significant because edge detection is an important capability in tactile sensing. In human tactile sensing edges are salient features \cite{plaisier_salient_2009}. However, most current approaches for tactile edge finding either rely on bespoke detection conditions or on methods from computer vision like \cite{platkiewicz_haptic_2016, lepora_pixels_2019}. These results indicate that our method is identifying some of the same high salience features as human touch without any explicit encoding of edges or corners in the exploration algorithm. 


In addition to overall lower real time training losses, we also observed that the entropy heat maps generated by the active learning method were crisper and more refined than their passive learning counterparts. This is particularly notable, because we found that after network convergence, the entropy heat maps serve to highlight important ``objects" in each tactile scene. This is particularly clear in Fig.~\ref{fig: leaf-n-figure} which shows entropy maps for both a leaf and the leather triangles. The placement and locations of high entropy areas are consistent between across training runs, suggesting that the entropy distribution itself can be used for object identification or localization. 

\section{Conclusion} 
Tactile sensing requires contact and measurements will vary depending on how a sensor is articulated. Here we present a method for generating exploratory motions without relying on primitives, hand crafting, or specialized knowledge of the objects under test. Our method produces lower losses when training perceptual tactile sensor models than a random walk baseline. We use an active learning approach in which the sensor spends more time in areas of low neural network certainty and less time in areas of high network certainty, avoiding prescribing haptic perception in terms of pre-formed features. The network entropy distributions also serve as a salience map of the tactile environment---including organic, difficult-to-model scenes---potentially providing an opportunity for salience-based registration and localization. 

We further show that without explicitly including edges and corners as haptic features, our method consistently finds edges and corners on simple scenes as an `emergent' byproduct of connecting exploration strategy to the learning model. This type of edge-centric exploratory motion is important in human tactile sensing for object recognition and may enable the same tactile capabilities in robots. Future work includes using these haptic models for object identification, registration, and localization.

\section*{Acknowledgments}

J.K. and T.M. acknowledge support from the Army Research Office (ARO, Grant No. W911NF-22-1-0286).
\balance
\clearpage

\bibliographystyle{IEEEtran}

\bibliography{references}

\end{document}